\documentclass[9pt,twocolumn,twoside]{pnas-new} 
\setboolean{displaywatermark}{false}

\templatetype{pnasresearcharticle} 
\providecommand{\scal}[2]{\left\langle{#1},{#2}\right\rangle}

\renewcommand{\O}{\mathcal{O}}

\DeclareMathOperator{\R}{\mathbb{R}}

\providecommand{\scal}[2]{\left\langle{#1},{#2}\right\rangle}
\newcommand{\be}{\begin{equation}}
\newcommand{\ee}{\end{equation}}
\newcommand{\bt}{\begin{theorem}}
\newcommand{\et}{\end{theorem}}
\newcommand{\bd}{\begin{definition}}
\newcommand{\ed}{\end{definition}}
\newcommand{\br}{\begin{remark}}
\newcommand{\er}{\end{remark}}

\renewcommand{\k}{\mathbf{k}}
\newcommand{\XX}{\mathbb{X}}
\newtheorem{property}{Property}
\newtheorem{definition}{Definition}
\newtheorem{conjecture}{Conjecture}

\newtheorem{proposition}{Proposition}
\newtheorem{remark}{Remark}

\newtheorem{theorem}{Theorem}
\newtheorem{fact}{Fact}

\title{Theoretical Issues in Deep
  Networks: Approximation, Optimization and
  Generalization}

\author[a,1]{Tomaso Poggio} 
\author[a]{Andrzej Banburski} 
\author[a]{Qianli Liao} 

\affil[a]{Center for Brains, Minds and Machines, MIT}

\leadauthor{Poggio} 

\significancestatement{ In the last few years, deep learning has been
  tremendously successful in many important applications of machine
  learning. However, our theoretical understanding of deep learning,
  and thus the ability of developing principled improvements, has
  lagged behind.  A theoretical characterization of deep learning is
  now beginning to emerge. It covers the following questions: 1) {\it
    representation power} of deep networks 2) {\it optimization} of
  the empirical risk 3) {\it generalization properties} of gradient
  descent techniques -- how can deep networks generalize despite being
  overparametrized -- more weights than training data -- in the
  absence of any explicit regularization? We review progress on all
  three areas showing that 1) for a the class of compositional
  functions deep networks of the convolutional type are exponentially
  better approximators than shallow networks; 2) only global minima
  are effectively found by stochastic gradient descent for
  over-parametrized networks; 3) there is a hidden norm control in the
  minimization of cross-entropy by gradient descent that allows
  generalization despite overparametrization. }

\authorcontributions{T.P. designed research; T.P., A.B., and Q.L. performed research; and T.P. and A.B. wrote the paper.}
\authordeclaration{The authors declare no conflict of interest.}
\correspondingauthor{\textsuperscript{1}To whom correspondence should be addressed. E-mail: tp@csail.mit.edu}


\begin{abstract}
  While deep learning is successful in a number of applications, it is
  not yet well understood theoretically.  A satisfactory theoretical
  characterization of deep learning however, is beginning to
  emerge. It covers the following questions: 1) {\it representation
    power} of deep networks 2) {\it optimization} of the empirical
  risk 3) {\it generalization properties} of gradient descent
  techniques --- why the expected error does not suffer, despite the
  absence of explicit regularization, when the networks are
  overparametrized? In this review we discuss recent advances in the
  three areas. In {\it approximation theory} both shallow and deep
  networks have been shown to approximate any continuous functions on
  a bounded domain at the expense of an exponential number of
  parameters (exponential in the dimensionality of the
  function). However, for a subset of compositional functions, deep
  networks of the convolutional type (even without weight sharing) can
  have a linear dependence on dimensionality, unlike shallow
  networks. In {\it optimization} we discuss the loss landscape for
  the exponential loss function. It turns out that global minima at
  infinity are completely degenerate. The other critical points of the
  gradient are less degenerate, with at least one -- and typically
  more -- nonzero eigenvalues.  This suggests that stochastic gradient
  descent will find with high probability the global minima. To
  address the question of {\it generalization} for classification
  tasks, we use classical uniform convergence results to justify
  minimizing a surrogate exponential-type loss function under a unit
  norm constraint on the weight matrix at each layer -- since the
  interesting variables for classification are the weight {\it
    directions} rather than the weights. As a
  side remark,  such minimization for
  (homogeneous) ReLU deep networks implies maximization of the
  margin. The resulting constrained gradient system turns out to be
  identical to the well-known {\it weight normalization} technique,
  originally motivated from a rather different way. We also show that
  standard gradient descent contains an implicit $L_2$ unit norm
  constraint in the sense that it solves the same constrained minimization
  problem with the same critical points (but a different dynamics).
  Our approach, which is supported by several
  independent new results \cite{theory_III,
    2017arXiv171010345S,DBLP:journals/corr/abs-1906-05890,2019arXiv190507325S},
  offers a solution to the puzzle about generalization performance of deep
  overparametrized ReLU networks, uncovering the origin of the
  underlying hidden complexity control in the case of deep networks.

\end{abstract}

\dates{This manuscript was compiled on \today}
\doi{\url{www.pnas.org/cgi/doi/10.1073/pnas.XXXXXXXXXX}}

\begin{document}

\maketitle
\thispagestyle{firststyle}
\ifthenelse{\boolean{shortarticle}}{\ifthenelse{\boolean{singlecolumn}}{\abscontentformatted}{\abscontent}}{}
\section{Introduction}
\dropcap{I}n the last few years, deep learning has been tremendously
successful in many important applications of machine
learning. However, our theoretical understanding of deep
learning, and thus the ability of developing principled
improvements, has lagged behind.  A satisfactory theoretical
characterization of deep learning is emerging. It covers the
following areas: 1) {\it approximation} properties of deep
networks 2) {\it optimization} of the empirical risk 3) {\it
	generalization} properties of gradient descent techniques
-- why the expected error does not suffer, despite the
absence of explicit regularization, when the networks are
overparametrized?

\subsection{When Can Deep  Networks Avoid the Curse of Dimensionality?} 

We start with the first set of questions,
summarizing results in \cite{HierarchicalKernels2015,Hierarchical2015,poggio2015December},
and \cite{Mhaskaretal2016,MhaskarPoggio2016}. The main result is that deep  networks have
the theoretical guarantee, which shallow networks do not have, that
they can avoid the {\it curse of dimensionality}  for an important class of problems,
corresponding to {\it compositional functions}, that is functions of
functions. An especially interesting subset of such compositional
functions are  {\it hierarchically local compositional
  functions} where all the constituent functions are local in the
sense of bounded small dimensionality. The deep networks that can
approximate them without the curse of dimensionality are of the deep
convolutional type -- though, importantly, weight sharing is not necessary. 

Implications of the theorems likely to be relevant in practice are:


 a) {\it Deep convolutional architectures} have the theoretical
  guarantee that they can be {\it  much better} than one layer architectures such
  as kernel machines for certain classes of problems;
 b) the problems for which certain deep networks are guaranteed to avoid
the {\it curse of dimensionality} (see for a nice review
  \cite{Donoho00high-dimensionaldata})  correspond to input-output
  mappings that are {\it compositional with local constituent
    functions};
c) the key aspect of convolutional networks that can give them an
  exponential advantage is {\it not weight sharing} but {\it locality} at each
  level of the hierarchy.


\subsection{Related Work}

Several papers in the '80s focused on the approximation power and
learning properties of one-hidden layer networks (called shallow
networks here). Very little appeared on multilayer networks, (but see
\cite{mhaskar1993approx, mhaskar1993neural, chui1994neural, chui1996, Pinkus1999}).
By now, several papers \cite{poggio03mathematics,MontufarBengio2014,
  DBLP:journals/corr/abs-1304-7045} have appeared. \cite{Anselmi2014,anselmi2015theoretical,poggioetal2015,
  LiaoPoggio2016, Mhaskaretal2016} derive new upper bounds for
the approximation by deep networks of certain important classes of
functions which avoid the curse of dimensionality. The upper bound for
the approximation by shallow networks of general functions was well
known to be exponential. It seems natural to assume that, since there
is no general way for shallow networks to exploit a compositional
prior, lower bounds for the approximation by shallow networks of
compositional functions should also be exponential. In fact, examples
of specific functions that cannot be represented efficiently by
shallow networks have been given,  for instance in
\cite{Telgarsky2015, SafranShamir2016, Theory_I}. An interesting
review of approximation of univariate functions by deep networks has
recently appeared \cite{2019arXiv190502199D}.

\begin{figure}
\centering
\includegraphics[trim=10 33 30 73, width=0.9\linewidth,clip]{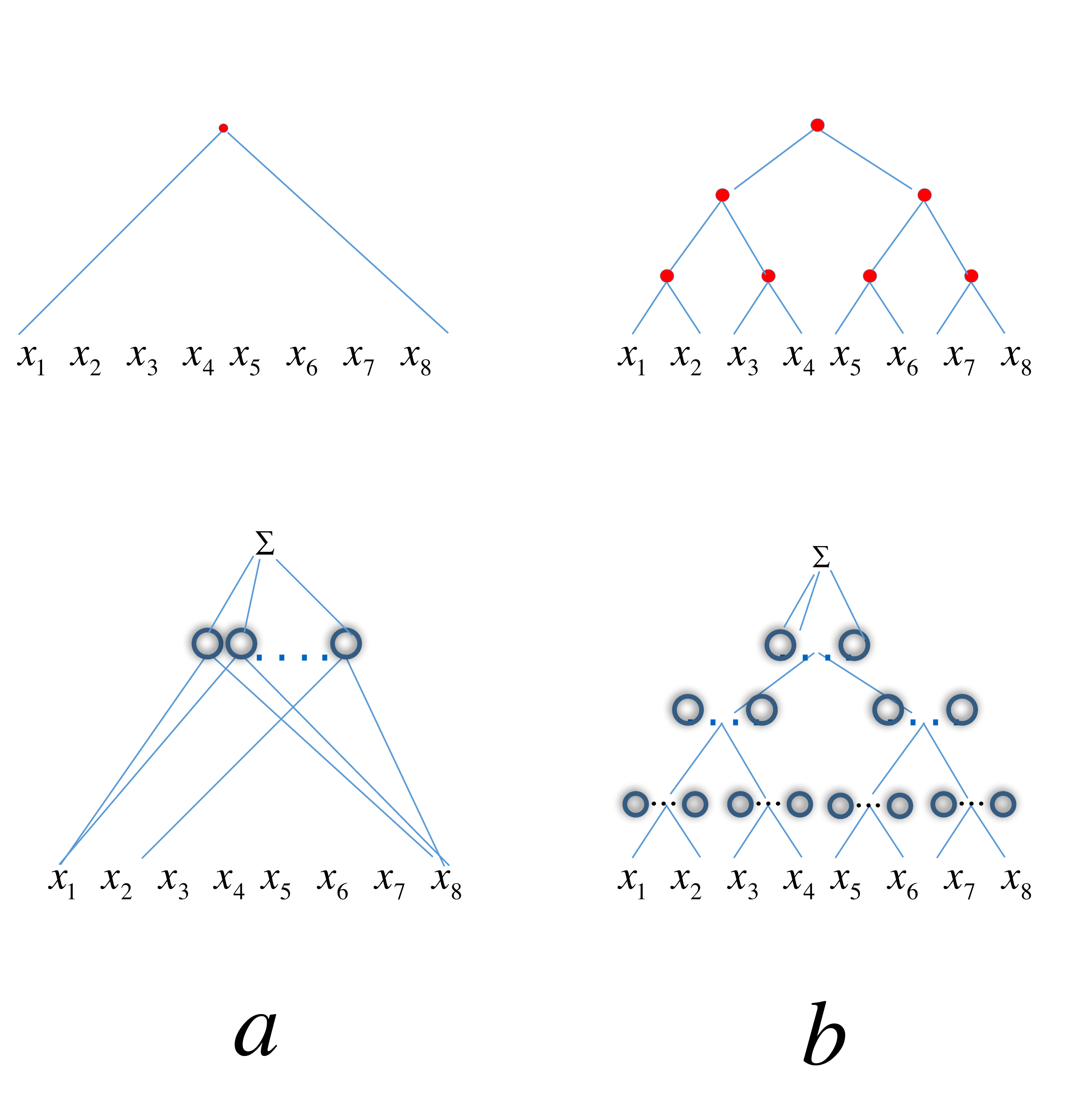}
\caption{The top graphs are associated to {\it functions}; each of the
  bottom diagrams depicts the ideal {\it network} approximating the
  function above. In a) a shallow universal network in 8 variables and
  $N$ units approximates a generic function of $8$ variables
  $f(x_1, \cdots, x_8)$. Inset b) shows a  hierarchical
  network at the bottom in $n=8$ variables, which approximates well
  functions of the form
  $f(x_1, \cdots, x_8) = h_3(h_{21}(h_{11} (x_1, x_2), h_{12}(x_3,
  x_4)), \allowbreak h_{22}(h_{13}(x_5, x_6), h_{14}(x_7, x_8))) $ as
  represented by the binary graph above.  In the approximating network
  each of the $n-1$ nodes in the graph of the function corresponds to
  a set of $Q =\frac{N}{n-1}$ ReLU units computing the ridge function
  $\sum_{i=1}^Q a_i(\scal{\mathbf{v}_i}{\mathbf{x}}+t_i)_+$, with
  $\mathbf{v}_i, \mathbf{x} \in \R^2$, $a_i, t_i\in\R$. Each term in
  the ridge function corresponds to a unit in the node (this is
  somewhat different from todays deep networks, but equivalent to
  them \cite{Theory_I}). Similar to the shallow network, a hierarchical network is
  universal, that is, it can approximate any continuous function; the
  text proves that it can approximate a compositional functions
  exponentially better than a shallow network. Redrawn from
  \cite{MhaskarPoggio2016}.
 }
\label{example_functions}
\end{figure}

\subsection{Degree of approximation}\label{approxsect}

The general paradigm is as follows. We are interested in determining
how complex a network ought to be to {\it theoretically guarantee}
approximation of an unknown target function $f$ up to a given accuracy
$\epsilon>0$.  To measure the accuracy, we need a norm $\|\cdot\|$ on
some normed linear space $\mathbb{X}$. As we will see the norm used in
the results of this paper is the $sup$ norm in keeping with the
standard choice in approximation theory. As it turns out, the results
of this section
require the sup norm in order to be independent from the unknown
distribution of the input data.

Let $V_N$ be the be set of all networks of a given kind with $N$ units
(which we take to be or measure of the complexity of the approximant
network). The \textit{degree of approximation} is defined by
$\mathsf{dist}(f, V_N)=\inf_{P\in V_N}\|f-P\|.$ For example, if
$\mathsf{dist}(f, V_N)=\O(N^{-\gamma})$ for some $\gamma>0$, then a
network with complexity $N=\O(\epsilon^{-\frac{1}{\gamma}})$ will be
sufficient to guarantee an approximation with accuracy at least
$\epsilon$. The only a priori information on the class of target
functions $f$, is codified by the statement that $f\in W$ for some
subspace $W\subseteq \mathbb{X}$. This subspace is a smoothness and compositional class,
characterized by the parameters $m$ and $d$ ($d=2$ in the example of
Figure \ref{example_functions} ; it is the size of the kernel
in a convolutional network).


\subsection{Shallow and deep networks}
\label{subprevious}

This section characterizes conditions under which deep networks are
``better'' than shallow network in approximating functions.  Thus we
compare shallow (one-hidden layer) networks with deep networks as
shown in Figure \ref{example_functions}.  Both types of networks use
the same small set of operations -- dot products, linear combinations,
a fixed nonlinear function of one variable, possibly convolution and
pooling. Each node in the networks   corresponds to
a node in the graph of the function to be approximated, as shown in
the Figure. A unit is a neuron which computes
$(\scal{x}{w}+b)_+$,  where $w$ is the vector of weights on the vector input
$x$. Both $w$ and the real number $b$ are parameters tuned by
learning. We assume here that each node in the networks computes the
linear combination of $r$ such units $\sum_{i=1}^r c_i
(\scal{x}{w_i}+b_i)_+$. Notice that in our main example of a network corresponding to a
function with a binary tree graph, the resulting architecture is an
idealized version of deep convolutional neural networks described in
the literature. In particular, it has only one output at the top
unlike most of the deep architectures with many channels and many
top-level outputs. Correspondingly, each node computes a single value
instead of multiple channels, using the combination of several
units. However our results hold also for these more
complex networks (see \cite{Theory_I}).
\noindent

The sequence of results is as follows.

\begin{itemize}

\item {\it Both shallow (a) and deep (b) networks are universal}, that
  is they can approximate arbitrarily well any continuous function of
  $n$ variables on a compact domain. The result for shallow networks
  is classical. 

\item We consider a special class of functions of $n$ variables on a
  compact domain that are {\it hierarchical compositions of local
    functions}, such as
  $f(x_1, \cdots, x_8) = h_3(h_{21}(h_{11} (x_1, x_2), h_{12}(x_3,
  x_4)), \allowbreak h_{22}(h_{13}(x_5, x_6), h_{14}(x_7, x_8))) $


\noindent The structure of the function in Figure  \ref{example_functions} b)
is represented by a graph of the binary tree type, reflecting dimensionality $d=2$ for the constituent functions $h$. In
general, $d$ is arbitrary but fixed and independent of the
dimensionality $n$ of the compositional function $f$.
\cite{Theory_I} formalizes the more general compositional case using
directed acyclic graphs.

\item The approximation of functions with a {\it compositional
    structure} -- can be achieved with the same degree of accuracy by
  deep and shallow networks but  the number of parameters are much
  smaller for the deep networks than for the shallow network with
  equivalent approximation accuracy.
\end{itemize}

We approximate functions with networks in which the activation
nonlinearity is a smoothed version of the so called ReLU, originally
called {\it ramp} by Breiman and given by
$\sigma (x) = x_+ = max(0, x)$ .  The architecture of the deep
networks reflects the function graph with each node $h_i$ being a
ridge function, comprising one or more neurons.

Let $I^n=[-1,1]^n$, $\XX=C(I^n)$ be the space of all continuous
functions on $I^n$, with $\|f\|=\max_{x\in I^n}|f(x)|$. 
Let
$\mathcal{S}_{N,n}$ denote the class of all shallow networks with $N$
units of the form
$$
x\mapsto\sum_{k=1}^N a_k\sigma(\scal{{w}_k}{x}+b_k),
$$
where ${w}_k\in\R^n$, $b_k, a_k\in\R$. The number of trainable
parameters here is $(n+2)N\sim n$. Let $m\ge 1$ be an integer, and
$W_m^n$ be the set of all functions of $n$ variables with continuous
partial derivatives of orders up to $m < \infty$ such that $\|f\|+\sum_{1\le
  |\k|_1\le m} \|D^\k f\| \le 1$, where $D^\k$ denotes the partial
derivative indicated by the multi-integer $\k\ge 1$, and $|\k|_1$ is
the sum of the components of $\k$. 

For the hierarchical binary tree network, the analogous spaces are
defined by considering the compact set $W_m^{n,2}$ to be the class of
all compositional functions $f$ of $n$ variables with a binary tree
architecture and constituent functions $h$ in $W_m ^2$.  We define the
corresponding class of deep networks $\mathcal{D}_{N,2}$ to be the set
of all deep networks with a binary tree architecture, where each of
the constituent nodes is in $\mathcal{S}_{M,2}$, where $N=|V|M$, $V$
being the set of non--leaf vertices of the tree. We note that in the
case when $n$ is an integer power of $2$, the total number of
parameters involved in a deep network in $\mathcal{D}_{N,2}$ is $4N$.

The first theorem is about shallow networks.

\begin{theorem}
\label{optneurtheo}
Let $\sigma :\R\to \R$ be infinitely differentiable, and not a polynomial.  For $f\in W_m^n$ the complexity of shallow networks that
provide accuracy at least $\epsilon$ is 
\be 
N= \O(\epsilon^{-n/m})\,\, and\,\, 
is\,\, the\,\, best\,\, possible.  
\ee
\end{theorem}

The estimate of Theorem \ref{optneurtheo} is the best possible if the only a priori
information we are allowed to assume is that the target function
belongs to $f\in W_m^n$. The exponential dependence on the
dimension $n$ of the number $e^{-n/m}$ of parameters needed to
obtain an accuracy $\O(\epsilon)$ is known as the {\it curse of
  dimensionality}. Note that the constants involved in $\O$ in the theorems will depend upon
the norms of the derivatives of $f$ as well as $\sigma$. 

Our second and main theorem is about deep networks with smooth
activations (preliminary versions appeared in
\cite{poggio2015December,Hierarchical2015,Mhaskaretal2016}). We
formulate it in the binary tree case for simplicity but it extends
immediately to functions that are compositions of constituent
functions of a fixed number of variables $d$ (in convolutional
networks $d$ corresponds to the size of the kernel).

\begin{theorem}
\label{deeptheo}
For $f\in W_m^{n,2}$ consider a deep network with the same
compositional architecture and with an activation function
$\sigma :\R\to \R$ which is infinitely differentiable, and
not a polynomial. The complexity of the
network to provide approximation with
accuracy at least $\epsilon$ is
\begin{equation}
N =\mathcal{O}((n-1)\epsilon^{-2/m}).
\label{deepnetapprox}
\end{equation}
\end{theorem}

The proof is in \cite{Theory_I}.  The assumptions on $\sigma$ in the
theorems are not satisfied by the ReLU function $x\mapsto x_+$, but
they are satisfied by smoothing the function in an arbitrarily small
interval around the origin. The result of the
theorem can be extended to non-smooth ReLU\cite{Theory_I}.

In summary, when the only a priori assumption on the target function
is about the number of derivatives, then to {\it guarantee} an
accuracy of $\epsilon$, we need a shallow network with
$\O(\epsilon^{-n/m})$ trainable parameters. If we assume a
hierarchical structure on the target function as in
Theorem~\ref{deeptheo}, then the corresponding deep network yields a
guaranteed accuracy of $\epsilon$ with $\O(\epsilon^{-2/m})$ trainable
parameters. Note that Theorem~\ref{deeptheo} applies to all $f$ with a
compositional architecture given by a graph which correspond to, or is
a subgraph of, the graph associated with the deep network -- in this
case the graph corresponding to $W_m^{n,d}$.

\section{ The Optimization Landscape of Deep
  Nets with Smooth Activation Function}
\label{BezoutBoltzman}

The main question in optimization of deep networks is to the landscape
of the empirical loss in terms of its global minima and local critical
points of the gradient. 

\subsection{Related work}

There are many recent papers studying optimization 
in deep learning. For optimization we mention work based on the idea
that noisy gradient descent \cite{DBLP:journals/corr/Jin0NKJ17,
  DBLP:journals/corr/GeHJY15, pmlr-v49-lee16, s.2018when} can find a
global minimum.  More recently, several authors studied the dynamics of
gradient descent for deep networks with assumptions about the input
distribution or on how the labels are generated. They obtain global
convergence for some shallow neural networks
\cite{Tian:2017:AFP:3305890.3306033, s8409482,
  Li:2017:CAT:3294771.3294828, DBLP:conf/icml/BrutzkusG17,
  pmlr-v80-du18b, DBLP:journals/corr/abs-1811-03804}. Some local
convergence results have also been proved
\cite{Zhong:2017:RGO:3305890.3306109,
  DBLP:journals/corr/abs-1711-03440, 2018arXiv180607808Z}. The most
interesting such approach is \cite{DBLP:journals/corr/abs-1811-03804},
which focuses on minimizing the training loss and proving that
randomly initialized gradient descent can achieve zero training loss
(see also \cite{NIPS2018_8038, du2018gradient,
  DBLP:journals/corr/abs-1811-08888}). In summary, there is by now an extensive
literature on optimization that formalizes and refines to different
special cases and to the discrete domain our results of  \cite{theory_II, theory_IIb}.

\subsection{Degeneracy of global and local minima under the
  exponential loss}

The {\it first part} of the argument of this section relies on the
obvious fact (see \cite{theory_III}), that for RELU networks under the
hypothesis of an exponential-type loss function, there are {\it no
  local minima that separate the data} -- the only critical points of
the gradient that separate the data are the global minima.

Notice that the global minima are at $\rho = \infty$, when the
exponential is zero. As a consequence, the Hessian is identically zero
with all eigenvalues being zero. On the other hand any point of the
loss at a finite $\rho$ has nonzero Hessian: for instance in the
linear case the Hessian is proportional to $\sum_n^N x_n x^T_n$.
The local minima which are not global minima  must
misclassify. How degenerate are they?

Simple arguments \cite{theory_III} suggest that the critical points
which are not global minima  cannot be completely degenerate. We thus have the following

\begin{property}
  Under the exponential loss, global minima are completely degenerate
  with all eigenvalues of the Hessian ($W$ of them with $W$ being the
  number of parameters in the network) being zero. The other critical
  points of the gradient are less degenerate, with at least one -- and typically $N$
  -- nonzero eigenvalues.
\end{property}

For the general case of non-exponential loss and smooth
nonlinearities instead of the RELU the following conjecture has been
proposed \cite{theory_III}:

\begin{conjecture}: For appropriate overparametrization, there are a
  large number of global zero-error minimizers which are degenerate;
  the other critical points -- saddles and local minima -- are
  generically (that is with probability one) degenerate on a set of
  much lower dimensionality.
\end{conjecture}

\subsection{SGD and Boltzmann Equation}
The second part of our argument (in \cite{theory_IIb}) is that SGD
concentrates in probability on the most degenerate minima. The argument is based on the
similarity between a Langevin equation  and SGD and on the fact
that the Boltzmann distribution is exactly the asymptotic ``solution'' of the
stochastic differential Langevin equation and also of SGDL, defined as
SGD with added white noise (see for instance
\cite{raginskyetal17}). The Boltzmann distribution is

\begin{equation}
p(f) = \frac{1}{Z}e^{-\frac{L}{T}},
\label{Bolzman}
\end{equation}

\noindent where $Z$ is a normalization constant, $L(f)$ is the loss
and $T$ reflects the noise power. The equation implies that SGDL
prefers degenerate minima relative to non-degenerate ones of the same
depth. In addition, among two minimum basins of equal depth, the one
with a larger volume is much more likely in high dimensions as shown
by the simulations in \cite{theory_IIb}. Taken together, these two
facts suggest that SGD selects degenerate minimizers corresponding to
larger isotropic flat regions of the loss. Then SDGL shows concentration --
{\it because of the high dimensionality} -- of its asymptotic
distribution Equation \ref{Bolzman}.

Together \cite{theory_II} and \cite{theory_III} suggest the following 

\begin{conjecture}: For appropriate
  overparametrization of the deep network, SGD selects with high
  probability the global minimizers of the empirical loss, which are
  highly degenerate.
\end{conjecture}


\begin{figure}
\centering
\includegraphics[width=1.0\linewidth ]{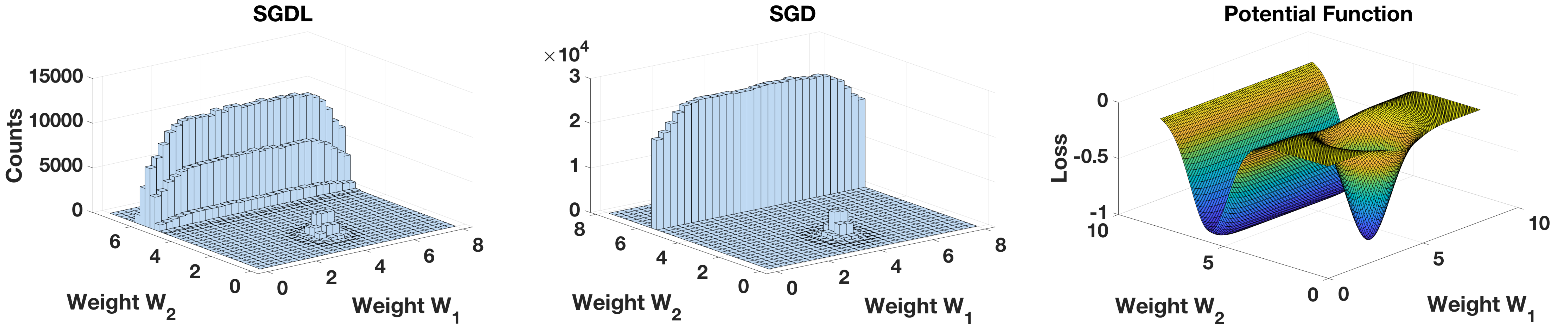}
\caption{ Stochastic Gradient Descent and Langevin Stochastic Gradient
  Descent (SGDL) on the $2$D potential function shown above leads to
  an asymptotic distribution with the histograms shown on the left. As
  expected from the form of the Boltzmann distribution, both dynamics
  prefer degenerate minima to non-degenerate minima of the same
  depth. From \cite{theory_III}.
}
\label{wedge_rbf_sgdl}
\end{figure}

\section{Generalization} 
\label{generalization}
\begin{figure}
	\centering
	\includegraphics[width=0.9\linewidth]{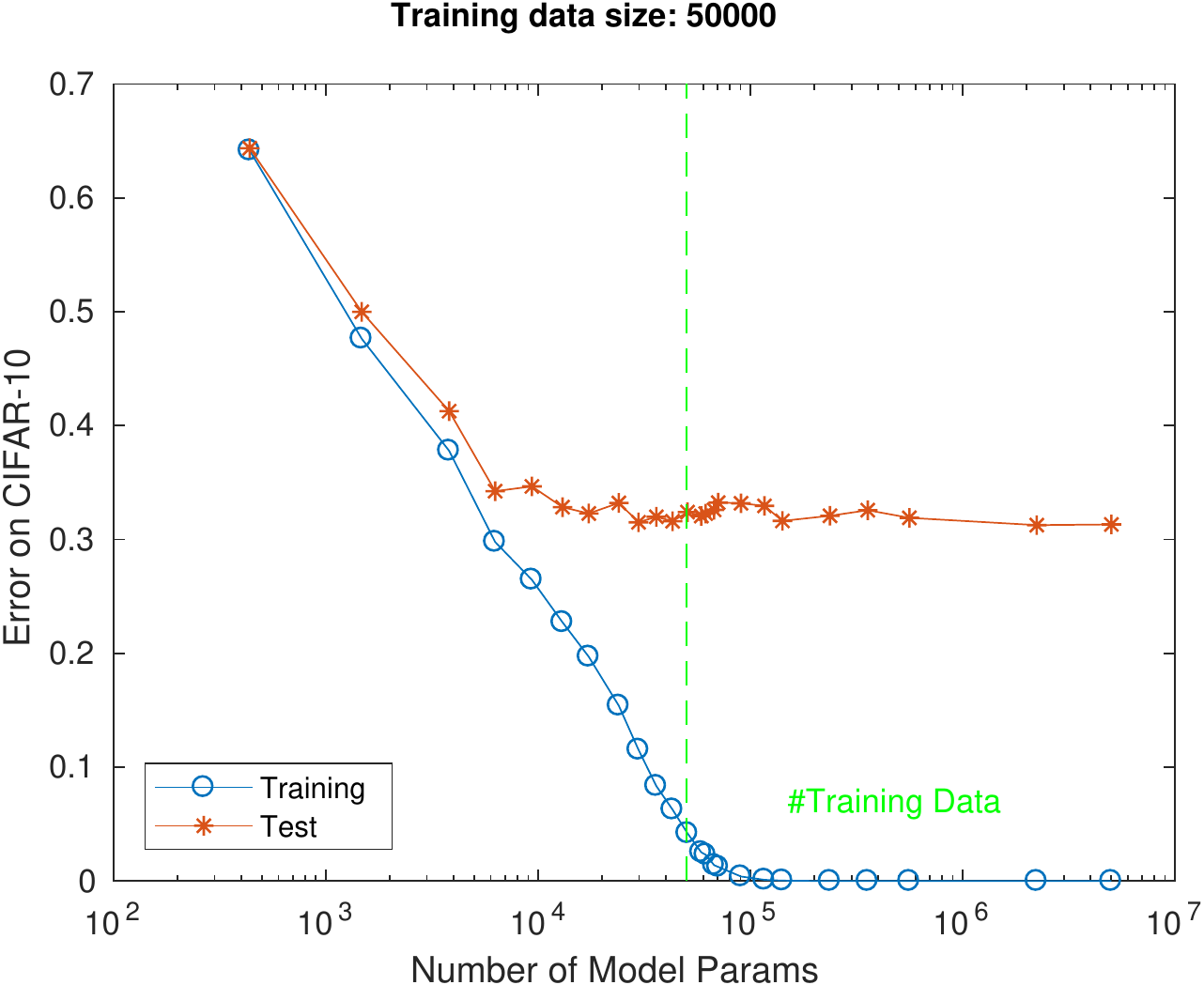}
	\caption{Empirical and expected error in CIFAR 10 as a
          function of number of neurons in a 5-layer convolutional
          network. The expected classification error does not increase
          when increasing the number of parameters beyond the size of
          the training set in the range we tested.}
	\label{no-overfitting}
\end{figure}

Recent results by \cite{2017arXiv171010345S} illuminate the apparent
absence of ''overfitting” (see Figure \ref{no-overfitting}) in the
special case of linear networks for binary classification.  They prove
that minimization of loss functions such as the logistic, the
cross-entropy and the exponential loss yields asymptotic convergence
to the maximum margin solution for linearly separable datasets,
independently of the initial conditions and without explicit
regularization.  Here we discuss the case of nonlinear multilayer DNNs
under exponential-type losses, for several variations of the basic
gradient descent algorithm. The main results are:
\begin{itemize}
\item classical uniform convergence bounds for generalization suggest
  a form of complexity control on the dynamics of the weight {\it
    directions $V_k$}: minimize a surrogate loss subject to a
  unit $L_p$ norm constraint;

\item gradient descent on the exponential loss with an explicit $L_2$
  unit norm constraint is equivalent to a well-known gradient descent
  algorithms {\it weight normalization} which is closely related to
  batch normalization;

\item unconstrained gradient descent on the exponential loss yields a
  dynamics with the same critical points as weight normalization: the
  dynamics implicitly respect a $L_2$ unit constraint on the
  directions of the weights $V_k$.
\end{itemize}

We observe that several of these results {\it  directly apply to kernel
machines} for the exponential loss under the 
separability/interpolation assumption, because  kernel machines are
one-homogeneous.

\subsection{Related work}

A number of papers have studied gradient descent for deep networks
\cite{NIPS2017_6836, DBLP:journals/corr/abs-1811-04918,
  Arora2019FineGrainedAO}. Close to the approach summarized here
(details are in \cite{theory_III}) is the paper
\cite{Wei2018OnTM}. Its authors study generalization assuming a
regularizer because they are -- like us -- interested in normalized
margin. Unlike their assumption of an explicit regularization, we show
here that commonly used techniques, such as weight and batch
normalization, in fact minimize the surrogate loss margin while
controlling the complexity of the classifier without the need to add a
regularizer or to use weight decay. Surprisingly, we will show that
even standard gradient descent on the weights implicitly controls the
complexity through an ``implicit'' unit $L_2$ norm constraint. Two
very recent papers (\cite{2019arXiv190507325S} and
\cite{DBLP:journals/corr/abs-1906-05890}) develop an elegant but
complicated margin maximization based approach which lead to some of
the same results of this section (and many more).  The important question of
which conditions are necessary for gradient descent to converge to the
maximum of the
margin of $\tilde{f}$ are studied by \cite{2019arXiv190507325S} and
\cite{DBLP:journals/corr/abs-1906-05890}. Our approach does not need the
notion of maximum margin but our theorem \ref{margin-maxTheorem}
establishes a connection with it and thus with the results of
\cite{2019arXiv190507325S} and
\cite{DBLP:journals/corr/abs-1906-05890}. Our main goal here (and in
\cite{theory_III}) is to achieve a simple understanding of where the
complexity control underlying generalization is hiding in the training
of deep networks.


\subsection{Deep networks: definitions and properties} 

We define a deep network with $K$ layers with the usual
coordinate-wise scalar activation functions
$\sigma(z):\quad \mathbf{R} \to \mathbf{R}$ as the set of functions
$f(W;x) = \sigma (W^K \sigma (W^{K-1} \cdots \sigma (W^1 x)))$, where
the input is $x \in \mathbf{R}^d$, the weights are given by the
matrices $W^k$, one per layer, with matching dimensions. We sometime
use the symbol $W$ as a shorthand for the set of $W^k$ matrices
$k=1,\cdots,K$. For simplicity we consider here the case of binary
classification in which $f$ takes scalar values, implying that the
last layer matrix $W^K$ is $W^K \in \mathbf{R}^{1,K_l}$. The labels
are $y_n\in\{-1,1\}$. The weights of hidden layer $l$ are collected in
a matrix of size $h_l\times h_{l-1}$.  There are no biases apart form
the input layer where the bias is instantiated by one of the input
dimensions being a constant. The activation function in this section
is the ReLU activation.

For ReLU activations the following
important positive one-homogeneity property holds
$\sigma(z)=\frac{\partial \sigma(z)}{\partial z} z$. 
A consequence of one-homogeneity is a structural lemma (Lemma 2.1 of
\cite{DBLP:journals/corr/abs-1711-01530}) $\sum_{i,j} W^{i,j}_k \left(\frac{\partial f(x)}{\partial
    W^{i,j}_k}\right)= f(x)$ where $W_k$ is here the vectorized representation of the weight
matrices $W_k$ for each of the different layers (each matrix is a
vector).

For the network, homogeneity implies
$f(W;x)=\prod_{k=1}^K \rho_k f(V_1,\cdots,V_K; x_n)$, where
$W_k=\rho_k V_k$ with the matrix norm $||V_k||_p=1$. Another property
of the Rademacher complexity of ReLU networks that follows from
homogeneity is
$\mathbb{R}_N(\mathbb{F}) = \rho \mathbb{R}_N(\tilde{\mathbb{F}})$
where $\rho=\rho_1 \prod_{k=1}^K \rho_k$, $\mathbb{F}$ is the class of
neural networks described above.

We define $f= \rho \tilde{f}$; $\tilde{\mathbb{F}}$ is the
associated class of normalized neural networks (we call
$f(V;x)=\tilde{f}(x)$ with the understanding that $f(x)=f(W;x)$).
Note that
$\frac{\partial f}{\partial \rho_k} = \frac{\rho}{\rho_k}\tilde{f}
\label{rho}$ and that the definitions of $\rho_k$, $V_k$ and
$\tilde{f}$ all depend on the choice of the norm used in
normalization.

In the case of training data that can be separated by the networks
$f(x_n) y_n>0 \quad \forall n=1,\cdots,N$. We will sometime write
$f(x_n)$ as a shorthand for $y_n f(x_n)$.

\subsection{Uniform convergence bounds: minimizing a surrogate loss
  under norm constraint}
\label{Early stopping}

Classical {\it generalization bounds for regression}
\cite{Bousquet2003} suggest that minimizing the empirical loss of a
loss function such as the cross-entropy
subject to constrained {\it complexity of the minimizer} is a way to
to attain  generalization, that is an expected loss which is close to the
empirical loss:

\begin{proposition}
The following generalization bounds 
apply to $\forall f \in \mathbb{F}$ with probability at least
$(1-\delta)$:
\begin{equation}
L(f) \leq \hat{L}(f) + c_1\mathbb{R}_N(\mathbb{F}) + c_2 \sqrt
\frac{\ln(\frac{1}{\delta})}{2N}
\label{bound}
\end{equation} 
\end{proposition}
\vskip0.1in
\noindent where $L(f) = \mathbf E [\ell(f(x), y)]$ is the expected
loss, $\hat{L}(f)$ is the empirical loss, $\mathbb{R}_N(\mathbb{F})$
is the empirical Rademacher average of the class of functions
$\mathbb{F}$, measuring its complexity; $c_1, c_2$ are constants that
depend on properties of the Lipschitz constant of the loss function,
and on the architecture of the network.

Thus minimizing under a constraint on the
Rademacher complexity  a surrogate function such as the
cross-entropy (which becomes the logistic loss in the binary
classification case) will minimize  an upper bound on the expected classification
error because such surrogate functions are upper bounds on the $0-1$
function. We can choose a class of functions $\mathbf{\tilde{F}}$ with
normalized weights and write $f(x)=\rho \tilde{f}(x)$ and
$\mathbb{R}_N(\mathbb{F})=\rho \mathbb{R}_N(\mathbb{\tilde{F}})$. One
can choose any fixed $\rho$ as a (Ivanov) regularization-type
tradeoff. 
In summary, the problem of generalization may be approached by minimizing
the exponential loss -- more in general an exponential-type loss, such
the logistic and the cross-entropy -- under a unit norm constraint on
the weight matrices, since we are interested in the directions of the weights:
\begin{equation}
\lim_{\rho \to \infty} \arg\min_{||V_k||=1, \  \forall k} L(\rho \tilde{f}) 
\label{UnitNormMin}
\end{equation} 
\noindent where we write $f(W) = \rho \tilde{f}(V)$ using the
homogeneity of the network. As it will become clear later, gradient
descent techniques on the exponential loss automatically increase
$\rho$ to infinity. We will typically consider the sequence of
minimizations over $V_k$ for a sequence of increasing $\rho$. The key
quantity for us is $\tilde{f}$ and the associated weights $V_k$;
$\rho$ is in a certain sense an auxiliary variable, a constraint that
is progressively relaxed.

In the following we explore the implications for deep networks of this classical
approach to generalization.

\subsubsection{Remark: minimization of an exponential-type loss implies margin maximization
}

Though not critical for our approach to the question of generalization
in deep networks it is interesting that constrained minimization of
the exponential loss implies margin maximization. This property
relates our approach to the results of several recent papers
\cite{2017arXiv171010345S,
  2019arXiv190507325S,DBLP:journals/corr/abs-1906-05890}. Notice that
our theorem \ref{margin-maxTheorem} as in
\cite{DBLP:conf/nips/RossetZH03} is a {\it sufficient condition for
  margin maximization}. Necessity is not true for general loss
functions.  

To state the margin property more formally, we adapt to our setting a
different result due to \cite{DBLP:conf/nips/RossetZH03} (they
consider for a linear network a vanishing $\lambda$ regularization
term whereas we have for nonlinear networks a set of unit norm
constraints). First we recall the definition of the empirical loss
$L(f)=\sum_{n=1}^N \ell(y_n f(x_n))$ with an exponential loss function
$\ell(yf)= e^{-yf}$.  We define $\eta(f)$ a the {\it margin} of $f$,
that is $\eta(f)=\min_n f(x_n)$.

Then our margin maximization theorem (proved in \cite{theory_III})  takes the form

\begin{theorem} 
    Consider the set of $V_k, k=1,\cdots, K$ corresponding to

  \begin{equation}
 \min_{{||V_k||}=1} L(f(\rho_k, V_k))
\label{V(rho)}
  \end{equation} 
  \noindent where the norm $||V_k||$ is a chosen $L_p$ norm and
  $L(f)(\rho_k, V_K) = L(\tilde{f}(\rho)) = \sum_ n \ell(y_n \rho f(V; x_n))$ is the
  empirical exponential loss. For each layer consider a sequence of increasing
  $\rho_k$. Then the associated sequence of $V_k$ defined by Equation
  \ref{V(rho)}, converges for $\rho \to \infty$ to the maximum
  margin of $\tilde{f}$, that is to
  $\max_{||V_k|| \leq 1} \eta(\tilde{f})$ .
\label{margin-maxTheorem}
\end{theorem}



\subsection{Minimization under unit norm constraint: weight normalization}
\label{OurWeightNormalization}

The approach is then to minimize the loss function
$ L(f(w))=\sum_{n=1}^N e^{- f(W;x_n) y_n }= \sum_{n=1}^N e^{- \rho
  f(V_k;x_n) y_n }$, with $\rho= \prod \rho_k$, subject to
$||V_k||^p_p =1 \ \forall k$, that is under a unit norm constraint for
the weight matrix at each layer (if $p=2$ then
$\sum_{i,j} (V_k)_{i,j}^2= 1$ is the Frobenius norm), since $V_k$ are
the directions of the weights which are the relevant quantity for
classification. The minimization is understood as a sequence of
minimizations for a sequence of increasing $\rho_k$. Clearly these
constraints imply the constraint on the norm of the product of weight
matrices for any $p$ norm (because any induced operator norm is a
sub-multiplicative matrix norm). The standard choice for a loss
function is an exponential-type loss such the cross-entropy, which for
binary classification becomes the logistic function. We study here the
exponential because it is simpler and retains all the basic
properties.

There are several gradient descent techniques that given the
unconstrained optimization problem transform it into a {\it
  constrained} gradient descent problem. To provide the background let us
formulate the standard unconstrained gradient descent problem for the
exponential loss as it is used in practical training of deep networks:

\begin{equation}
	\dot{W}^{i,j}_k  = -\frac{\partial L}{\partial W^{i,j}_k}= \sum_{n=1}^N y_n \frac{\partial{f(x_n; w)}} {\partial W^{i,j}_k} e^{- y_n
		f(x_n;W)} 
\label{standardynamicsW}
\end{equation}
\noindent where $W_k$ is the weight matrix of layer $k$. Notice that,
since the structural property implies that at a critical point we have
$\sum_{n=1}^N y_n f(x_n; w) e^{- y_n f(x_n;W)} = 0$, the only critical
points of this dynamics that separate the data (i.e.
$y_n f(x_n; w)>0 \ \forall n$) are global minima at infinity. Of
course for separable data, while the loss decreases asymptotically to
zero, the norm of the weights $\rho_k$ increases to infinity, as we
will  see later. Equations \ref{standardynamicsW} define a dynamical
system in terms of the gradient of the exponential loss $L$.

The set of gradient-based algorithms enforcing a unit-norm constraints
\cite{845952} comprises several techniques that are equivalent for small
values of the step size. They are all good approximations of the true
gradient method. One of them is the {\it Lagrange multiplier method};
another is the {\it tangent gradient method} based on the following theorem:

\begin{theorem} \cite{845952}  Let $||u||_p$ denote a vector norm that is
  differentiable with respect to the elements of $u$ and let $g(t)$ be
  any vector function with finite $L_2$ norm.  Then, calling
  $\nu(t)=\frac{\partial ||u||_p}{\partial u}_{u=u(t)}$, the equation
\begin{equation}
 \dot{u}=h_g(t)=Sg(t)= (I-\frac{\nu \nu^T}{||\nu||_2^2}) g(t)
\label{dot_u}
\end{equation}
\noindent with $||u(0)|| =1$, describes the flow of a vector $u$ that
satisfies $||u(t)||_p=1$ for all $t \ge 0$. 
\label{Theorem1}
\end{theorem}

In particular, a form for $g$ is $g(t)= \mu(t) \nabla_u L$, the
gradient update in a gradient descent algorithm. We call $Sg(t)$ the
tangent gradient transformation of $g$.  In the case of $p=2$ we replace $\nu$ in Equation
\ref{dot_u} with $u$ because
$\nu(t)=\frac{\partial ||u||_2}{\partial u}=u$. This gives
$S= I-\frac{u u^T}{||u||_2^2}$ and $\dot{u}=Sg(t).$

Consider now the empirical loss $L$ written in terms of $V_k$ and
$\rho_k$ instead of $W_k$, using the change of variables defined by
$W_k=\rho_k V_k$ but without imposing a unit norm constraint on $V_k$.
The flows in $\rho_k,V_k$ can be computed as
$\dot{\rho_k}=\frac{\partial W_k}{\partial \rho_k} \frac{\partial
  L}{\partial W_k}= V_k^T \frac{\partial L}{\partial W_k}$ and
$\dot{V_k}=\frac{\partial W_k}{\partial V_k} \frac{\partial L}{\partial W_k} =
\rho_k \frac{\partial L}{\partial W_k}$, with $\frac{\partial
  L}{\partial W_k}$ given by Equations \ref{standardynamicsW}.

We now enforce the unit norm constraint on $V_k$ by using the tangent gradient
transform on the $V_k$ flow. This yields

\begin{equation}
\dot{\rho_k}= V_k^T \frac{\partial  L}{\partial W_k} \quad
\dot{V_k}= S_k \rho_k \frac{\partial L}{\partial W_k}.
\label{v-flow-withunitnorm}
\end{equation}

Notice that the dynamics above follows from the classical approach of
controlling the Rademacher complexity of $\tilde{f}$ during optimization (suggested
by bounds such as Equation \ref{bound}.  The approach and the
resulting dynamics for the directions of the weights may seem  different from the standard
unconstrained approach in training deep networks. It turns out,
however, that the dynamics described by Equations
\ref{v-flow-withunitnorm} is the same dynamics of {\it Weight
  Normalization}.

The technique of {\it Weight normalization} \cite{SalDied16} was
originally proposed as a small improvement on standard gradient descent
``to reduce covariate shifts''. It was defined for each layer in
terms of $w=g \frac{v}{||v||}$, as
\begin{equation}
\dot{g}=\frac{v}{||v||} \frac{\partial L}{\partial w}
\dot{v}=\frac{g}{||v||} S \frac{\partial L}{\partial w}
\label{W-normalization}
\end{equation}
\noindent with $S=I- \frac{v v^T}{||v||^2}$.  

It is easy to see that Equations \ref{v-flow-withunitnorm} are the
same as the weight normalization Equations \ref{W-normalization}, if
$||v||_2=1$. We now observe, multiplying Equation
\ref{v-flow-withunitnorm} by $v^T$, that $v^T \dot{v}=0$ because
$v^T S=0$, implying that $||v||^2$ is constant in time with a constant
that can be taken to be $1$. Thus the two dynamics are the
same. 



\subsection{Generalization with hidden
  complexity control}
\label{nounitnorm}
Empirically it appears that GD and SGD converge to solutions that can
generalize even without batch or weight normalization.  Convergence
may be difficult for quite deep networks and generalization may not be
as good as with batch normalization but it still occurs. How is this
possible?  

We study the dynamical system 
$\dot{W_k}^{i,j}$ under the reparametrization
$W^{i,j}_k = \rho_k V^{i,j}_k$ with $||V_k||_2=1$.  We consider  for each
weight matrix $W_k$ the corresponding ``vectorized'' representation in
terms of vectors $W_k^{i,j} = W_k$.  We use the following definitions
and properties (for a vector $w$):

\begin{itemize}
\item Define $\frac{w}{||w||_2}=\tilde{w}$; thus $w=||w||_2\tilde{w}$ with
  $||\tilde{w}||_2=1$. Also define $S={I-\tilde{w}\tilde{w}^T}=I- \frac {w
      w^T}{||w||_2^2}$.
\item The following relations are easy to check:
\begin{enumerate}
\item $\frac{\partial ||w||_2}{\partial w}=\tilde{w}$
\item $\frac{\partial \tilde{w}}{\partial
    w}=\frac{S}{||w||_2}$. 
\item $Sw=S \tilde{w}=0$
\item $S^2=S$
\label{Relations}
\end{enumerate}
\end{itemize}

The gradient descent dynamic system used in training deep networks for
the exponential loss is given by Equation \ref{standardynamicsW}.
Following the chain rule {\it for the time derivatives}, the dynamics
for $W_k$ is exactly (see \cite{theory_III}) equivalent to the
following dynamics for $||W_k||=\rho_k$ and $V_k$:
\begin{equation}
\dot{\rho_k}= \frac{\partial ||W_k||}{\partial W_k} \frac{\partial
  W_k}{\partial t}= V^T_k \dot{W_k}
\label{rhodot}
\end{equation}
\noindent and
\begin{equation}
\dot{V_k}= \frac{\partial V_k}{\partial W_k} \frac{\partial
  W_k}{\partial t}= \frac {S_k}{\rho_k} \dot{W_k}
\label{vdot}
\end{equation}
\noindent where $S_k= I- V_kV_k^T$.  We used property 1 in
\ref{Relations} for Equation \ref{rhodot} and property 2 for Equation
\ref{vdot}.
%
%

The key point here is that the dynamics of $\dot{V_k}$ includes a unit
$L_2$ norm constraint: using the tangent gradient transform will not
change the equation because $S^2=S$.

As separate remarks , notice that if for $t>t_0$, $f$ separates all
the data, $\frac{d}{dt}{\rho_k} >0$, that is $\rho$ diverges to
$\infty$ with $\lim_{t \to \infty}\dot{\rho}=0$. In the 1-layer
network case the dynamics yields $\rho \approx \log t$
asymptotically. For deeper networks, this is
different. \cite{theory_III} shows (for one support vector) that the
product of weights at each layer diverges faster than logarithmically,
but each individual layer diverges slower than in the 1-layer case.
The norm of the each layer grows at the same rate $\dot{\rho_k^2}$,
independent of $k$. The $V_k$ dynamics has stationary or critical points given by

\begin{equation}
\sum \alpha_n(\rho(t)
\left(\frac{\partial{\tilde{f}(x_n)}} {\partial
                  V_k^{i,j}}-V_k^{i,j}  \tilde{f}(x_n) \right),
\label{wdot4}
\end{equation}
\noindent where $\alpha_n= e^{-y_n \rho(t) \tilde{f}(x_n)}$. 
We examine later the linear one-layer case $\tilde{f}(x)=v^T x$ in
which case the stationary points of the gradient are given by $\sum \alpha_n(\rho(t)
(x_n - v v^T x_n)$ and of course coincide with the solutions obtained
with Lagrange multipliers. In the general case the
critical points correspond for $\rho \to \infty$ to degenerate zero
``asymptotic minima'' of the loss.


To understand whether there exists an implicit complexity control in
standard gradient descent of the weight directions, we  check whether there exists an
$L_p$ norm for which unconstrained normalization is equivalent to
constrained normalization.

From Theorem \ref{Theorem1} we expect
the constrained case to be given by the action of the following
projector onto the tangent space:
\begin{equation}
  S_{p} = I-\frac{\nu \nu^T}{||\nu||_2^2} \quad\textnormal{with}\quad \nu_i=\frac{\partial ||w||_p}{\partial w_i} = \textnormal{sign}(w_i)\circ\left(\frac{|w_i|}{||w||_p}\right)^{p-1}.   
\end{equation}
The constrained Gradient Descent is then
\begin{equation}
\dot{\rho_k}= V^T_k \dot{W_k}  \quad
\dot{V_k} = \rho_k S_p \dot{W_k}.
\label{ConstrainedGradP}
\end{equation}

On the other hand,  reparametrization of
the unconstrained dynamics in the $p$-norm gives (following Equations \ref{rhodot} and \ref{vdot})
\begin{equation}
\begin{split}
\dot{\rho_k}&= \frac{\partial ||W_k||_p}{\partial W_k} \frac{\partial
	W_k}{\partial t}= \textnormal{sign}(W_k)\circ\left(\frac{|W_k|}{||W_k||_p}\right)^{p-1} \cdot \dot{W_k} \\
\dot{V_k}&= \frac{\partial V_k}{\partial W_k} \frac{\partial
	W_k}{\partial t}= \frac {I - \textnormal{sign}(W_k) \circ\left(\frac{|W_k|}{||W_k||_p}\right)^{p-1}W_k^T}{||W_k||_p^{p-1}} \dot{W_k}.
\end{split}
\end{equation}
These two dynamical systems are clearly different for generic
$p$ reflecting the presence or absence of a regularization-like
constraint on the dynamics of $V_k$.

As we have seen however, for $p=2$ the 1-layer dynamical system obtained by
minimizing $L$ in $\rho_k$ and $V_k$ with $W_k=\rho_k V_k$ under the constraint
$||V_k||_2=1$, is the weight normalization dynamics

\begin{equation}
\dot{\rho_k}=V_k^T \dot{W_k} \quad
\dot{V_k}= S \rho_k \dot{W_k} ,
\label{WN}
\end{equation}

\noindent which is quite similar to the standard gradient
equations 
\begin{equation}
\dot{\rho_k}= V_k^T \dot{W_k}  \quad
\dot{v} =\frac{S}{\rho_k} \dot{W_k}.
\label{StandardGrad}
\end{equation}

\begin{figure*}[t!]\centering
		\includegraphics[width=1.0\textwidth]{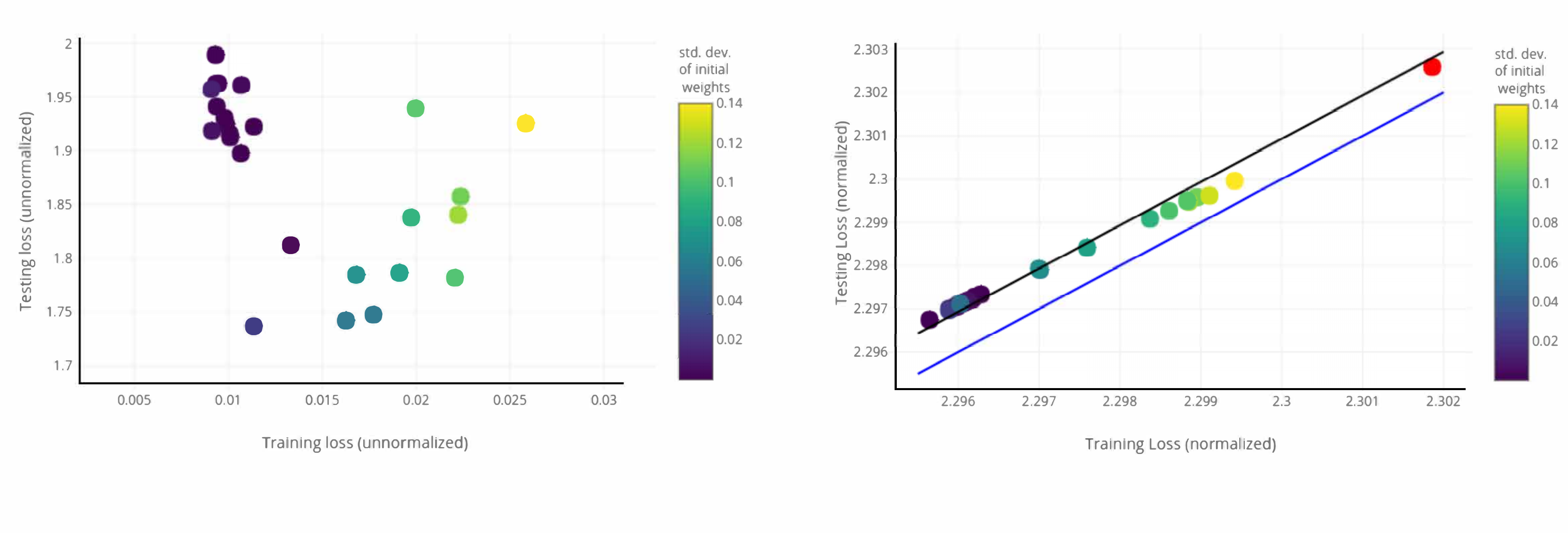} 
		\caption{\it The top left graph shows testing vs
                  training cross-entropy loss for networks each
                  trained on the same data sets (CIFAR10) but with a
                  different initializations, yielding zero
                  classification error on training set but different
                  testing errors. The top right graph shows the same
                  data, that is testing vs training loss for the same
                  networks, now normalized by dividing each weight by
                  the Frobenius norm of its layer. Notice that all
                  points have zero classification error at
                  training. The red point on the top right refers to a
                  network trained on the same CIFAR-10 data set but
                  with randomized labels. It shows zero classification
                  error at training and test error at chance
                  level. The top line is a square-loss regression of
                  slope $1$ with positive intercept. The bottom line
                  is the diagonal at which training and test loss are
                  equal.  The networks are 3-layer convolutional
                  networks. The left can be considered as a
                  visualization of Equation \ref{bound} when the
                  Rademacher complexity is not controlled. The right
                  hand side is a visualization of the same relation
                  for normalized networks that is
                  $L(\tilde{f}) \leq \hat{L}(\tilde{f}) +
                  c_1\mathbb{R}_N(\mathbb{\tilde{F}}) + c_2 \sqrt
                  \frac{\ln(\frac{1}{\delta})}{2N}$. Under our
                  conditions for $N$ and for the architecture of the
                  network the terms
                  $c_1\mathbb{R}_N(\mathbb{\tilde{F}}) + c_2 \sqrt
                  \frac{\ln(\frac{1}{\delta})}{2N}$ represent a small
                  offset. From
                  \cite{DBLP:journals/corr/abs-1807-09659}.
}
\label{main}
	\end{figure*}

        The two dynamical systems differ only by a $\rho_k^2$ factor
        in the $\dot{V_k}$ equations. However, the critical points of
        the gradient for the $V_k$ flow, that is the point for which
        $\dot{V_k}=0$, are the same in both cases since for any $t>0$
        $\rho_k(t)>0$ and thus $\dot{V_k}=0$ is equivalent to
        $S\dot{W_k}=0$.  Hence, gradient descent with unit $L_p$-norm
        constraint is equivalent to the standard, unconstrained
        gradient descent but only when $p = 2$. Thus

        \begin{fact} The standard dynamical system used in deep
          learning, defined by
          $\dot{W_k}=-\frac{\partial L}{\partial W_k}$, implicitly
          respectss a unit $L_2$ norm constraint on $V_k$ with
          $\rho_k V_k =W_k$. Thus, under an exponential loss, if the
          dynamics converges, the $V_k$ represent the minimizer under
          the $L_2$ unit norm constraint.
	\label{w(T)}
\end{fact}

Thus standard GD implicitly enforces the $L_2$ norm constraint on
$V_k=\frac{W_k}{||W_k||_2}$, consistently with Srebro's results on
implicit bias of GD. Other minimization techniques such as coordinate
descent may be biased towards different norm constraints.

\subsection{Linear networks and rates of convergence}

The linear ($f(x)=\rho v^T x$) networks case
\cite{2017arXiv171010345S} is an interesting example of our analysis
in terms of $\rho$ and $v$ dynamics. We start with unconstrained
gradient descent, that is with the dynamical system

 \begin{equation}
\dot{\rho}=  \frac{1}{\rho} \sum_{n=1}^N  e^{-
                \rho v^Tx_n} v^Tx_n \quad \dot{v}=\frac{1}{\rho}\sum_{n=1}^N  e^{- \rho v^T x_n} 
(x_n- v  v^T x_n).
\label{Ttt}
\end{equation}

If gradient descent in $v$ converges to $\dot{v}=0$ at finite time,
$v$ satisfies $ v v^T x = x$, where $x= \sum_{j=1}^C \alpha_j x_j$
with positive coefficients $\alpha_j$ and $x_j$ are the $C$ support
vectors (see \cite{theory_III}).  A solution $v^T = ||x|| x^\dagger$
then {\it exists} ($x^\dagger$, the pseudoinverse of $x$, since $x$ is
a vector, is given by $x^\dagger= \frac{x^T}{||x||^2}$). On the other
hand, the operator $T$ in $v(t+1)=T v(t)$ associated with equation
\ref{Ttt} is non-expanding, because $||v||=1,\ \forall t$. Thus there
is a fixed point $v \propto x$ which is {\it independent of initial
  conditions} \cite{Ferreira1996} and unique (in the linear case)

The rates of convergence of the solutions $\rho(t)$ and $v(t)$,
derived in different way in \cite{2017arXiv171010345S}, may be read
out from the equations for $\rho$ and $v$. It is easy to check that a
general solution for $\rho$ is of the form $\rho \propto C \log t$. A
similar estimate for the exponential term gives
$e^{- \rho v^T x_n} \propto \frac{1}{t}$. Assume for simplicity a
single support vector $x$. We claim that a solution for
the error $\epsilon= v-x$, since $v$ converges to $x$, behaves as
$\frac{1}{\log t}$.  In fact we write $v =x+\epsilon$ and plug it in
the equation for $v$ in \ref{T}. We obtain (assuming normalized input $||x||=1$)
\begin{equation}
\dot{\epsilon}=\frac{1}{\rho}  e^{- \rho v^T x} (x- (x+\epsilon)
(x+\epsilon)^T x) \approx \frac{1}{\rho}  e^{- \rho v^T x} ( x- x - x
\epsilon^T - \epsilon x^T),
\label{T}
\end{equation}
\noindent which has the form
$\dot{\epsilon}=-\frac{1}{t \log t} (2 x \epsilon^T)$. Assuming
$\epsilon$ of the form
$\epsilon \propto \frac{1}{\log t}$ we obtain
$-\frac{1}{t \log^2 t }= -B \frac{1}{t \log^2 t}$. Thus the error
indeed converges as
$\epsilon \propto \frac{1}{\log t}$.

A similar analysis for the weight normalization equations \ref{WN}
considers the same dynamical system with a change in the equation for
$v$, which becomes

\begin{equation}
\dot{v} \propto e^{-\rho}  \rho  (I- v v^T) x.
\label{T-WN}
\end{equation}
This equation differs by a factor $\rho^2$ from
equation \ref{T}. As a consequence equation \ref{T-WN} is of the form
$\dot{\epsilon}=-\frac{\log t}{t} \epsilon$, with a general solution
of the form $\epsilon \propto t^{-\frac{1}{2}\log t}$. In
summary, {\it GD with weight normalization converges faster to the
  same equilibrium than standard gradient descent: the rate for
  $\epsilon= v- x$ is  $t^{-\frac{1}{2} log(t)}$ vs  $\frac{1}{\log t}$.} 
 
Our goal was to find
$\lim_{\rho \to \infty} \arg \min_{||V_k||=1, \ \forall k} L(\rho
\tilde{f}) $. We have seen that various forms of gradient descent
enforce different paths in increasing $\rho$ that empirically have
different effects on convergence rate. It will be an interesting
theoretical and practical challenge to find the optimal way, in terms
of generalization and convergence rate, to grow $\rho\rightarrow \infty$.

Our analysis of simplified batch normalization \cite{theory_III}
suggests that several of the same considerations that we used for
weight normalization should apply (in the linear one layer case BN is
identical to WN). However, BN differs from WN in the multilayer case
in several ways, in addition to weight normalization: it has for
instance separate normalization for each unit, that is for each row of
the weight matrix at each layer.

\section{Discussion}

A main difference between shallow and deep networks is in terms of
{\it approximation} power or, in equivalent words, of the ability to
learn good representations from data based on the compositional
structure of certain tasks. Unlike shallow networks, deep local
networks -- in particular convolutional networks -- can avoid the
curse of dimensionality in approximating the class of hierarchically
local compositional functions. This means that for such class of
functions deep local networks represent an appropriate hypothesis
class that allows good approximation with a minimum number of
parameters. It is not clear, of course, why many problems encountered
in practice should match the class of compositional functions. Though
we and others have argued that the explanation may be in either the
physics or the neuroscience of the brain, these arguments are not
rigorous. Our conjecture at present is that compositionality is
imposed by the wiring of our cortex and, critically, is reflected in
language. Thus compositionality of some of the most common visual
tasks may simply reflect the way our brain works.

{\it Optimization} turns out to be surprisingly easy to perform for
overparametrized deep networks because SGD will converge with high
probability to global minima that are typically much more degenerate for
the exponential loss than other local critical points.

More surprisingly, gradient descent yields {\it generalization} in
classification performance, despite overparametrization and even in
the absence of explicit norm control or regularization, because
standard gradient descent in the weights enforces an implicit unit
($L_2$) norm constraint on the {\it directions of the weights} in the
case of exponential-type losses.

In summary, it is tempting to conclude that the practical success of
deep learning has its roots in the almost magic synergy of unexpected
and elegant theoretical properties of several aspects of the
technique: the deep convolutional network architecture itself, its
overparametrization, the use of stochastic gradient descent, the
exponential loss, the homogeneity of the RELU units and of the
resulting networks.

Of course many problems remain open on the way to develop a full
theory and, especially, in translating it to new architectures. More
detailed results are needed in approximation theory, especially for
densely connected networks. Our framework for optimization is missing
at present a full classification of local minima and their dependence
on overparametrization for general loss functions. The analysis of
generalization should include an analysis of convergence of the
weights for multilayer networks (see \cite{2019arXiv190507325S} and
\cite{DBLP:journals/corr/abs-1906-05890}). A full theory would also
require an analysis of the trade-off  between
approximation and estimation error, relaxing the separability
assumption.

%
%

\showmatmethods{} 

\acknow{We are grateful to Sasha Rakhlin and Nate Srebro for useful
  suggestions about the structural lemma and about separating critical
  points. Part of the funding is from the Center for Brains, Minds and
  Machines (CBMM), funded by NSF STC award CCF-1231216, and part by
  C-BRIC, one of six centers in JUMP, a Semiconductor Research
  Corporation (SRC) program sponsored by DARPA.}

\showacknow{} 

\bibliography{Boolean}

\end{document}